\documentclass{article}

\usepackage[final]{neurips_2019}

\usepackage[utf8]{inputenc} 
\usepackage[T1]{fontenc}    
\usepackage{hyperref} 
\usepackage[nameinlink, capitalise]{cleveref}
\usepackage{url}      
\usepackage{booktabs} 
\usepackage{amsfonts} 
\usepackage{nicefrac} 
\usepackage{microtype}      
\usepackage{bm}
\usepackage{algorithm, algorithmic}    
\usepackage{amsmath}
\usepackage{amssymb}
\usepackage{makecell}
\usepackage{array}
\usepackage{graphicx}
\usepackage{tabularx}
\usepackage{siunitx}
\usepackage{stackrel}
\usepackage{multirow}

\usepackage[table]{xcolor}
\usepackage{colortbl}
\usepackage{multirow}
\usepackage{numprint}

\floatname{algorithm}{Algorithm}

\DeclareMathOperator*{\argmin}{arg\,min}

\newcommand{\MSuccessRate}{\textbf{SR}}
\newcommand{\MMean}{\textbf{Mean}}
\newcommand{\MWeightedMean}{\textbf{WMean}}

\newcommand{\MMeanDistOrigin}{\textbf{$\text{MD}_\text{O}$}}
\newcommand{\MWeightedMeanDistOrigin}{\textbf{$\text{WMD}_\text{O}$}}

\title{Imperceptible Adversarial Attacks on Tabular Data}

\author{
 Vincent~Ballet$^{*\dag}$\quad Xavier~Renard$^\dag$\quad Jonathan~Aigrain$^\dag$\\
  \textbf{Thibault~Laugel$^{\ddagger}$\quad Pascal~Frossard$^*$\quad Marcin~Detyniecki$^{\dag\ddagger\S}$} \\
  $^*$École Polytechnique Fédérale de Lausanne\\
  $^\dag$AXA, Paris, France\\
  $^\ddagger$Sorbonne Université, CNRS, LIP6, Paris, France\\
  $^\S$Polish Academy of Science, IBS PAN, Warsaw, Poland\\
  \href{mailto:vincent.ballet@me.com}{\texttt{vincent.ballet@me.com}}, \href{mailto:pascal.frossard@epfl.ch}{\texttt{pascal.frossard@epfl.ch}}, \href{mailto:thibault.laugel@lip6.fr}{\texttt{thibault.laugel@lip6.fr}}\\
  \texttt{\{}\href{mailto:xavier.renard@axa.com}{\texttt{xavier.renard}}, \href{mailto:jonathan.aigrain@axa.com}{\texttt{jonathan.aigrain}}, \href{mailto:marcin.detyniecki@axa.com}{\texttt{marcin.detyniecki}}\texttt{\}@axa.com}
}

\begin{document}

\maketitle

\begin{abstract}
    Security of machine learning models is a concern as they may face adversarial attacks for unwarranted advantageous decisions. While research on the topic has mainly been focusing on the image domain, numerous industrial applications, in particular in finance, rely on standard tabular data. In this paper, we discuss the notion of adversarial examples in the tabular domain. We propose a formalization based on the imperceptibility of attacks in the tabular domain leading to an approach to generate imperceptible adversarial examples. Experiments show that we can generate imperceptible adversarial examples with a high fooling rate.
\end{abstract}

\section{Introduction}\label{sec:intro}



As machine learning becomes more prevalent in many industries, concerns are raised about the security its usgae. Research has shown that machine learning models are sensitive to slight changes in their input, resulting in unwarranted predictions, application failures or errors.
The Adversarial Machine Learning field studies this issue with the design of attacks and defenses with an active focus on the image domain~\citep{43405, evasionattacksbiggio, advexamples}.
However, many machine learning classifiers in the industry rely on tabular data in input to predict labels in a wide variety of tasks (e.g. stocks, credit, fraud, etc.).

\begin{figure}[t]
 \centering
  \includegraphics[width=0.8\linewidth]{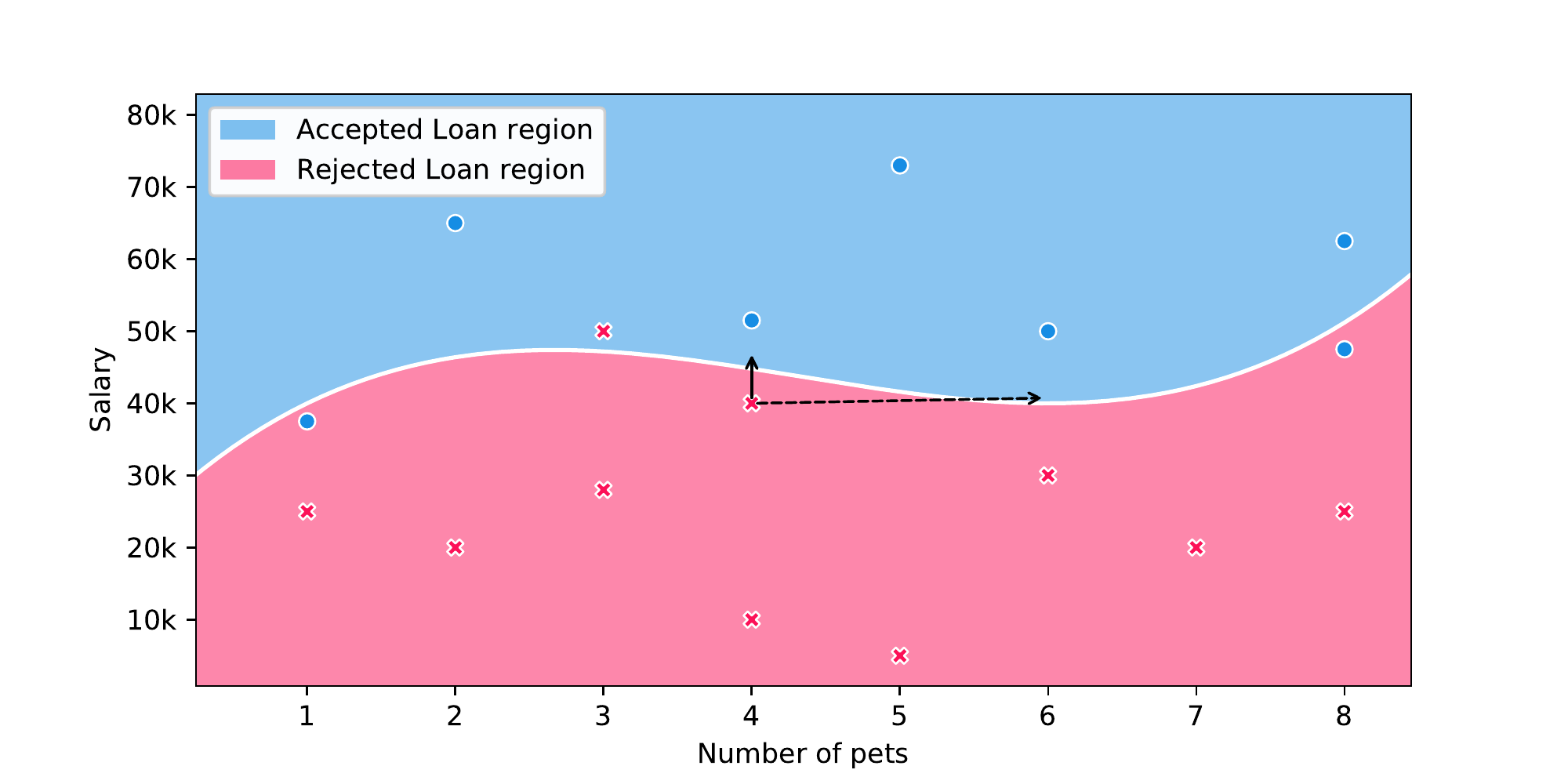}
  \caption{Illustration of an imperceptible perturbation to craft an adversarial attack on tabular data. The plain arrow is a "naive", perceptible, perturbation to cross the decision boundary (e.g. salary for a loan application). The dashed arrow is an imperceptible perturbation relying on the number of pets.}
  \label{fig:intuition}
\end{figure}

To illustrate the threat of adversarial attacks in a tabular context, we consider the scenario where a bank customer applies for a loan. A machine learning model is used to make a decision regarding the acceptance of the application based on customer provided information (incomes, age, etc.). The model advises the bank to reject the application of our customer who is determined to get the loan by filling false information to mislead the model.
In this work, we claim that the key for this attack to succeed is its imperceptibility: the application should remain credible and relevant for a potential expert eye, in coherence with the model's prediction.
We discuss the notions of imperceptibility and relevance in order to design an attack relying on the intuition that only a subset of features are critical for the prediction according to the expert eye (e.g. income, age). Thus, the attacker should minimize manipulations on this subset for the attack to be imperceptible and instead rely on less important features to get the model to accept the application. The intuition behind this idea is depicted Figure~\ref{fig:intuition}.

In the following section, we discuss the notion of \emph{imperceptibility} in order to propose a definition for adversarial examples in the context of tabular data.
Using this formalization, we propose in~\cref{sec:generate} an approach to generate adversarial examples called LowProFool.
To assess these propositions, we perform experiments on four literature datasets. We show in \cref{sec:expe} and \cref{sec:discussion} that the generated adversarial examples have a low perceptibility compared to other adversarial attacks from the state-of-the-art while the success rate (fooling rate) for the attack is kept high.

\section{Imperceptible adversarial attacks on images} \label{sec:impimages}

Defining the exact properties of an adversarial example remains a subject of debate among the research community, in particular in terms of \textit{imperceptibility} of the attack. On the one hand, \citeauthor{defadvcarlini} define the adversarial example as inputs that are close to \textit{natural} inputs but classified incorrectly~\citep{defadvcarlini} and~\citeauthor{szegedyintruiguing} mention \textit{imperceptible} non-random perturbation~\citep{szegedyintruiguing}. On the other hand, \citeauthor{locapatch} allow the noise to be \textit{visible} but confined to a small, localized patch of the image~\citep{locapatch} and~\citeauthor{advpatch} generate a \textit{discernible} adversarial patch that is stamped on the original picture to make the classifier always predict the same class~\citep{advpatch}.

While on images it is natural for a human to check whether the model has been fooled by comparing the image and the output of the classifier, assessing or measuring the perceptibility of the perturbation is complex. The most commonly used measure is the $\ell_p$ norm. For instance, ~\citeauthor{szegedyintruiguing} formalizes the definition of adversarial example and \textit{slight} perturbation in the image domain as 
$
    f(\bm x + \bm r) \neq f(\bm x) \text{ with } ||\bm r||_{p} < \epsilon
$
where $f$ is a classifier, $\bm x$ an image and $\bm r$ the adversarial perturbation. The \textit{slight} perturbation is controlled by $\epsilon$, the upper bound of its amplitude.

However, it is sometimes inaccurate to use the $\ell_p$ norm to assess the perceptibility of a perturbation. Evidence of that came to light recently when~\citeauthor{lpnorm} showed that the $\ell_p$ norm is insufficient to measure perceptual similarity between two images. The authors studied how images that are really close in terms of human-observed perceptibility (e.g. rotated versions of the same image) can still have large differences according to the $\ell_p$ norm. On the contrary, some images may be semantically different but considered close by this metric~\citep{lpnorm}.

Given these studies on the imperceptibility of attacks on images, in the next section we consider the specificities of tabular data to formalize the notion of imperceptibility for this context.


\section{Formalization of the notion of imperceptible attacks on tabular data}
\label{sec:formalization}

The notion of imperceptibility and the way it can be measured differ for tabular data compared to images for two main reasons. First, tabular features are not interchangeable like pixels. Second, while most people can usually tell the correct class of an image and whether it appears altered or not, it is much complex for tabular data: this type of data is less readable and expert knowledge is required.
On tabular data, to decide if two instances share the same class, the expert is likely to focus on a subset of features (among the whole feature space) he considers important for the classification task (eg. to predict the acquisition of a loan, incomes are more important than other features).
Thus, to detect the presence of any fraudulent modification, the expert is likely to check more thoroughly the veracity of this subset of features: the attacker should avoid modifications on it.
Then, we propose to measure the perceptibility of an adversarial attack as the $\ell_p$ norm of the perturbation weighted by a feature importance vector, which describes the likelihood of a feature to be investigated by the expert.
Given that tabular data cannot be manipulated the same way as images, we argue that using the $\ell_p$ norm is not subject to the same extent to the flaws discussed in~\cref{sec:impimages}.

Another aspect of the attack that is intrinsically contained in the notion of imperceptibility is the coherence of the output. While adversarial examples for image data naturally satisfy this constraint (pixels are defined as integers between $0$ and $255$), this needs to be enforced for tabular data. In fact, we expect the generated attacks to lie in natural, intuitive or hard bounds designed by the human intuition or the knowledge of an expert. If the adversarial example does not satisfy these constraints, the perturbed features must be bounded and rounded so as to fit into their context: for instance, the age might be forced into a positive number, or a boolean to be discrete depending on the problem.

To craft adversarial examples, we make the assumption that perturbations on less relevant features allow to reach the desired opposed class label. In fact, we seek to exploit the discrepancies between the classifier's learned vector of feature importance and the expert's knowledge. The perturbation is expected to be of a higher $l_p$ norm compared to optimal shortest paths to the classifier frontier. Despite being of higher norm, the perturbation should be less perceptible than another perturbation of smaller norm but supported by more relevant features for the experts.

Formally, we consider a set of examples $\mathbb{X}$ where each example is denoted by $\bm x^{(i)}$ with $i \in [1...N]$ and associated with a label $y^{(i)}$. The set of examples $\mathbb{X}$ is defined by a set of features $j \in \mathbb{J}$ and each feature vector is noted $\bm x_j$ with $j \in \mathbb{J}=[1...D]$. Also we consider $f: \mathbb R^D \rightarrow \{0, 1\}$, a binary classifier mapping examples to a discrete label $l \in \{0, 1\}$ and $d: \mathbb R^D \rightarrow [0, 1]$ a mapping between the perturbation $\bm r \in \mathbb R^D$ and its perceptibility value. Finally, we define $A \subseteq \mathbb R^D$ the set of valid, coherent samples.

For a given $\bm x \in \mathbb R^D$, its original label $s = f(\bm x)$ and a target label $t \neq s$, we aim at generating the optimal perturbation vector $\bm r^* \in \mathbb R^D$ such that

\begin{equation} \label{eq:argmin_d}
\begin{gathered}
\bm r^* = \argmin_{\bm r} d(\bm r) \quad \text{for } \bm r \in \mathbb R^D\\
\text{s.t.} \quad f(\bm x) = s \neq f(\bm x+ \bm r^*)=t \quad \text{and} \quad \bm x+ \bm r^* \in A
\end{gathered}
\end{equation}


As mentioned earlier, each feature $j \in \mathbb{J}$ is associated to a feature importance coefficient $v_j \in \mathbb{R}$ gathered in a feature importance vector
$
    \bm v = [v_1, \dots, v_j, \dots, v_{D}] \quad \text{where} \quad v_j > 0,\quad \forall j \in \mathbb{J}
$

We extend the definition of $d$ to include the feature importance $\bm v$. The perceptibility for tabular data is then defined as the $\ell_p$ norm of the feature importance weighted perturbation vector, such that:
\begin{equation}
    d_{\bm v}(\bm r) = ||\bm r \odot \bm v||_p^2 \quad \text{where} \odot \text{is the Hadamard product}
\end{equation}

Finally, in Equation~\ref{eq:argmin_d}, the constraint $\bm x+ \bm r^* \in A$ brings the idea of coherence of the generated adversarial example. The nature of regular tabular data requires that methods respect the discrete, categorical and continuous features. Each feature must conform with the dataset so as to be as imperceptible as possible, e.g. the feature \textit{age} should not be lower than 0.



\section{Generation of Imperceptible Adversarial Examples on Tabular Data} \label{sec:generate}

Our objective is to craft adversarial examples such that perturbations applied to a tabular data example are imperceptible to the expert's eye, i.e. higher perturbations for irrelevant or less important features, than for relevant ones, as described in Section~\ref{sec:formalization}.

Since solving the proposed minimization problem is complex, we propose to define the objective function as an aggregation of the class change constraint and the  minimization of $d_v$: $$g(\bm r) = \mathcal L (\bm x  + \bm r, t) + \lambda ||\bm v \odot  \bm r||_{p}$$

With $\mathcal{L}(x, t)$ the value of the loss of the model $f$ calculated for $x$ and target class $t$, and $\lambda > 0$. On the one hand, the regularizer $||\bm v \odot \bm r||_{p}$ allows to minimize the perturbation $\bm r$ with respect to its perceptibility. 
On the other hand, by minimizing the loss $\mathcal L(\bm x + \bm r, t)$, we ensure that the perturbation leads the perturbed sample towards the target class. 

Setting the value of the hyperparameter $\lambda \in \mathbb R$ allows to control the weight associated to the penalty of using important features, with respect to the feature importance vector. Given a fixed number of iterations, our goal is to minimize the weighted norm $||\bm v \odot  \bm r||_{p}$ associated with each adversarial example, while maximizing the proportion of samples $\bm x \in \mathbb X$ that could cross the classification frontier. These two optimization problems constitute a trade-off that is represented by choosing the right value for $\lambda$.

The LowProFool (low profile) algorithm is then defined as an optimization problem in which we search for the minimum of the objective function $g$ using a gradient descent approach.
More concretely, we make use of the gradient data so as to guide the perturbation towards the target class in an iterative manner. At the same time, we penalize the perturbation proportionally to the feature importance associated with the features, so as to minimize the perceptibility of the adversarial perturbation as defined in \cref{sec:formalization}.
\cref{alg:1} outline LowProFool algorithm for binary classifiers and features in the continuous domain.




\begin{algorithm}[t]  \label{sec:implem}
    \caption{LowProFool}
    \label{alg:1}
    \begin{algorithmic}[1]
  \REQUIRE Classifier $f$, sample $\bm x$, target label $t$, loss function $\mathcal L$, trade-off factor $\lambda$, feature importance $\bm v$, maximum number of iterations $N$, scaling factor $\alpha$
  \ENSURE Adversarial example $\bm x'$
      \STATE $\bm r \leftarrow [0,0,\dots, 0]$
      \STATE $\bm x_0 \leftarrow \bm x$
      \FOR{$i$ in $0 \dots N-1$} 
  
   \STATE $\bm r_i \leftarrow - \nabla_{\bm r} ( \mathcal L (\bm x_i, t) + \lambda ||\bm v \odot \bm r||_{p})$
   \STATE $\bm r \leftarrow \bm r + \alpha  \bm  r_i$
   \STATE $\bm x_{i+1} \leftarrow \text{clip}(\bm x + \bm r)$
       \ENDFOR
      \STATE $\bm x' \leftarrow \argmin_{\bm x_i} d_{\bm v}(\bm x_i) \quad \forall i \in [0 \dots N-1] \quad\text{s.t.}\quad f(\bm x_i) \neq f(\bm x_0)$
    \RETURN $\bm x'$
    \end{algorithmic}
\end{algorithm}

\section{Experimentation framework}\label{sec:expe}

\subsection{Metrics for adversarial attacks on tabular data} \label{sec:metrics}

\paragraph{Success Rate : } \label{sec:metricsSR}
The success rate or fooling rate is a common metric to measure the efficiency of an adversarial attack. Let us define the set $\hat{\mathbb X}$ that comprises every tuple $(\bm x, \bm x')$ such that $\bm x \in \mathbb X$ and $\bm x'$ is a successfully crafted adversarial example from $\bm x$, that is: $f(\bm x) \neq f(\bm x')$.

For a given number of iterations $N$ of~\cref{alg:1}, we then define the success rate $\sigma_N$ as: $$\sigma_N = \cfrac{|\hat{\mathbb X}|}{|\mathbb X|}$$

\paragraph{Norms of perturbation : }  \label{sec:metricsNorm}
To evaluate how successful an attack is, we also measure the norm of the adversarial perturbation.
$$||\bm r||_p$$

However, as discussed in~\cref{sec:formalization}, we measure the perceptibility of the perturbation by calculating the weighted norm $d_{\mathbf v}(\mathbf{r})$.

For both weighted and non-weighted perturbation norms we compute the mean value per pair of datasets and method over the set of samples $\mathbb X$.

\paragraph{Distance to the closest neighbor : } \label{sec:metricsNeigh}

Although perturbation norms allow us to compare methods between one another, they do not give insight about how significant the perturbation is for a given dataset. To this end, we propose to compute the average weighted distance to the closest neighbor of each original sample $\bm x$. This metric helps us develop an intuition about what is the distance between two points in the dataset and how it compares to the norm of the perturbation.

Formally, for a sample $\bm x \in \mathbb R^D$ we compute its closest neighbor as the following $$\text{neigh}(\bm x) = \argmin_{\bm p \in \mathbb X} d_{\bm v}(\bm x - \bm p)$$

\subsection{Data} 

\paragraph{Datasets : }We run experiments on four well-known datasets: German Credit dataset~\citep{Dua:2019}, Australian Credit dataset~\citep{Dua:2019}, the Default Credit Card dataset~\citep{defaultcredit} and the Lending Club Loan dataset~\citep{lendingloan}. They are all related to the financial service industry, hence representing a good case study for the scenario we considered in \cref{sec:intro}.

\paragraph{Preprocessing : } \label{sec:preproc}
The proposed notion applies to numerical continous data. Hence, in our experiments discrete features are treated as continous and non-ordered categorical features are dropped. 

Each dataset is split into train, test and validation sets. Test and validation subsets respectively hold 250 and 50 samples. All the remaining samples are used to train the model.
The test data is used to generate the adversarial examples and the validation data used to optimize hyperparameters.

\subsection{Parameters of the adversarial attack}

\paragraph{Objective function : }
As described in~\cref{sec:generate}, the objective function is defined as $g(\bm r) = \mathcal L (\bm x  + \bm r, t) + \lambda ||\bm v \odot \bm r||_{p}$. We defined the loss function $\mathcal L$ as the the binary cross entropy function and use the $\ell_2$ norm to optimize the weighted norm $||\bm v \odot  \bm r||_{2}$

\paragraph{Feature importance :}
In order to model the expert's knowledge, we use the absolute value of the Pearson's correlation coefficient of each feature with the target variable $t$.
We scale the obtained feature importance vector $\bm v$ by multiplying it by the inverse of its $\ell_2$ norm so as to feed the algorithm with a unit vector, irrespective of the dataset at hand.
More formally, $$\bm v = \cfrac{|\bm \rho_{X,Y}|}{ ||\bm \rho_{X,Y}||_2^2} \quad \text{where }|\bm u| = [|u_0|, |u_1|, \dots]$$

It is important to note that the expert's knowledge could be modelled using various other definitions. Pearson's correlation was chosen for its simplicity and its potential to replicate the intuition of an expert through a linear correlation at the scale of the dataset.

\paragraph{Clipping :}
To make sure that the generated adversarial examples $\bm x'$ lie in a coherent subspace $A \subseteq \mathbb R^D$, we clip $\bm x'$ to the bounds of each feature. More formally we constrain each feature $j \in [1\dots D]$ to stay in their natural range $[\min(\bm x_j), \max(\bm x_j)]$.

\paragraph{Machine Learning Model :} 
For each dataset we build a fully connected neural network using dense ReLU layers and a Softmax layer for the last one.

\paragraph{Comparison with other attacks:}
To the best of our knowledge, no method has been proposed to generate adversarial examples of tabular data. We hence compare LowProFool against two other methods that are FGSM~\citep{43405} and DeepFool~\citep{deepfool}, both being state-of-the-art baselines for gradient-based methods in the image domain. To do so, we use the metrics defined in \cref{sec:metrics}.

\subsection{Experimental results}
We report in~\cref{table:results} the results for the metrics defined in~\cref{sec:metrics}. The mean perturbation (\MMean), weighted mean perturbation (\MWeightedMean) and distance to the closest neighbors (\MMeanDistOrigin) as well as its weighted version (\MWeightedMeanDistOrigin) are computed only for pairs of samples $(\bm x, \bm x')$ such that $\bm x$ is a sample that belongs to the test set $\mathbb X$ and $\bm x'$ is an adversarial example crafted from $\bm x$ that successfully crosses the classifier's frontier, i.e. $f(\bm x) \neq f(\bm x')$. This allows us to compare the methods between each other.

We observe that LowProFool succeeds in fooling the classifier (\MSuccessRate) almost $95\%$ of the time. Only one dataset shows lower performances with a success rate of 86\%. DeepFool's success rate is about $100\%$ on each dataset while FGSM performs very poorly, from $0\%$ to $59\%$ success rate.

Then, comparing the mean $\ell_2$ perturbation norm of DeepFool and LowProFool shows that DeepFool always leads to better results. Averaging on the four datasets, the perturbation norm of LowProFool is $1.7$ times higher that DeepFool's. FGSM shows order of magnitude larger perturbation norms that the two other methods.

\begin{figure}[t]
 \centering
  \includegraphics[width=0.8\linewidth]{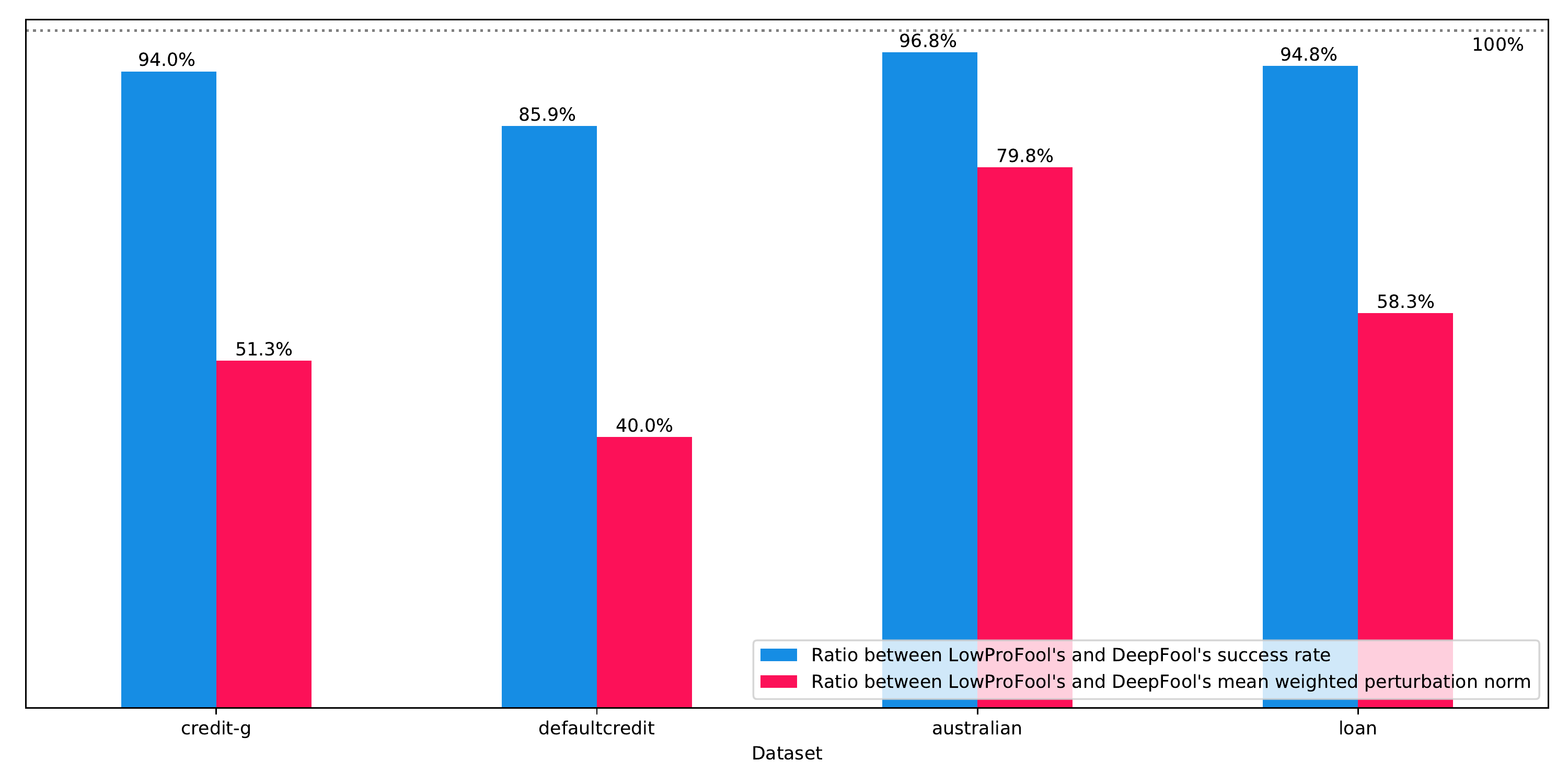}
  \caption{Comparison between state-of-the-art DeepFool and LowProFool. The blue bars show the ratio between the success rate of LowProFool and DeepFool. The red bars exhibit the ratio between LowProFool mean weighted perturbation norm and DeepFool mean weighted perturbation norm.}
  \label{fig:overall}
\end{figure}

Looking at the weighted perturbation, we observe that DeepFool consistently leads to worst results in comparison with LowProFool. We even get a ratio of 60\% between the weighted mean norm of LowProFool and DeepFool on the Default Credit Card dataset. Figure~\ref{fig:overall} is a diagrammatic comparison of discrete results presented in \cref{table:results} and allows better visual comparison between DeepFool and LowProFool.

\begin{figure}[H]
 \centering
  \includegraphics[width=0.8\linewidth]{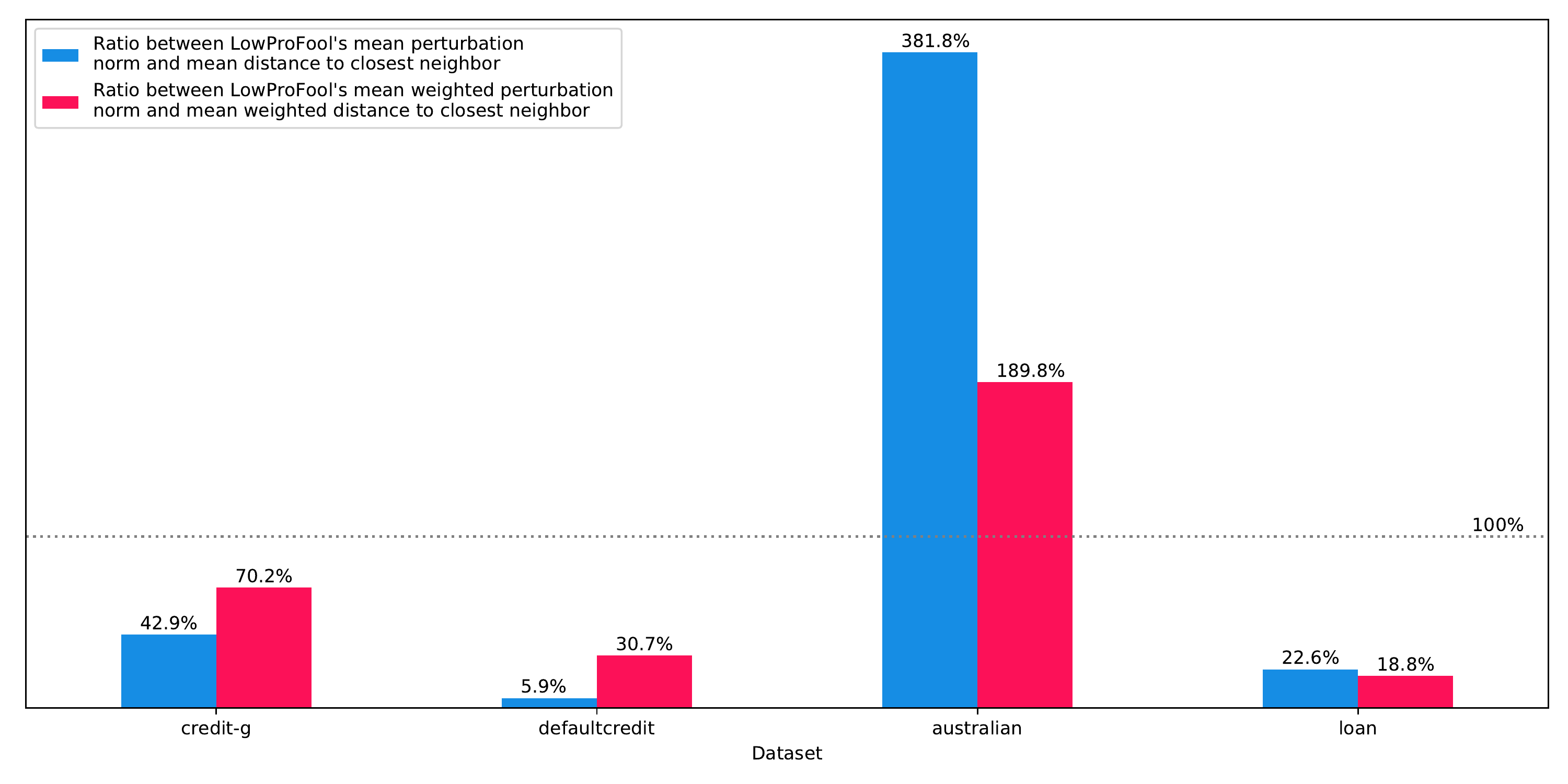}
  \caption{Comparison between the perturbation norm and the distance to the closest neighbor. Results are shown as ratio between the mean (weighted) perturbation norm and the mean (weighted) distance to the closest neighbor.}
  \label{fig:distneigh}
\end{figure}

To get a feeling of what the mean perturbation norms represent for each dataset, i.e. what order of magnitude are the perturbations, we compare them to the mean distance between each original sample and their closest neighbor in terms of weighted (\MWeightedMeanDistOrigin) and non-weighted (\MMeanDistOrigin) distance. We observe in Figure~\ref{fig:distneigh} that except for the Australian Credit dataset, the mean perturbation norm is in average $60\%$ smaller than the mean distance to the closest neighbors and the weighted mean perturbation norm $77\%$ smaller than the mean weighted distance to the closest neighbors.
On the contrary, we observe that it takes a higher perturbation for adversarial examples to be generated on the Australian Credit dataset. In fact in terms of weighted distance, the perturbation is four times bigger than the distance to the closest neighbor. This relates with \cref{table:results} in which we observe that the Australian Credit dataset shows higher perturbation norms and distance to neighbors than any other dataset studied.

\begin{center}
\begin{table}[t]
\small
\centering
  \caption{Results of the tests ran on the test set for each dataset.~\MSuccessRate~is the success rate defined in \cref{sec:metricsSR}.~\MMean~and~\MWeightedMean~respectively correspond to the mean norm of the perturbation and its weighted equivalent as defined in \cref{sec:metricsNorm}. \MMeanDistOrigin~and \MWeightedMeanDistOrigin~refer to the distance to the closest at the original sample, based on a $\ell_2$ distance and its weighted version $d_{\bm v}$ as proposed in \cref{sec:metricsNeigh}}\label{table:results}
\begin{tabularx}{\textwidth}{|c|l|l|X|X|X|X|}
\hline
\rotatebox[origin=c]{90}{\textbf{\hspace{1ex}Dataset\hspace{1ex}}} & \textbf{Method} & \MSuccessRate &  \MMean  & \MWeightedMean & \MMeanDistOrigin  & \MWeightedMeanDistOrigin \\ 
\hline
\hline

\multirow{3}{*}[-0.5ex]{\rotatebox[origin=c]{90}{German C.}}&\textbf{LowProFool}   &0.94 & 0.344 $\pm$ 0.282 & \textbf{0.039 $\pm$ 0.027} & 0.49 $\pm$ 0.193 & 0.091 $\pm$ 0.039\\
\cline{2-7}
& \textbf{DeepFool} & 1.0 & 0.21 $\pm$ 0.181 & 0.076 $\pm$ 0.077  & 0.485 $\pm$ 0.193 & 0.089 $\pm$ 0.038 \\
\cline{2-7}
& \textbf{FGSM}   &  0.192 & 19.69 $\pm$ 75.61 & 5.683 $\pm$ 21.827 & 0.477 $\pm$ 0.173 & 0.085 $\pm$ 0.037\\
\hline
\hline
\multirow{3}{*}[-0.5ex]{\rotatebox[origin=c]{90}{Default C.}} & \textbf{LowProFool} & 0.856 & 0.061 $\pm$ 0.109 & \textbf{0.002 $\pm$ 0.005} & 0.199 $\pm$ 0.137 & 0.034 $\pm$ 0.031 \\
\cline{2-7}
&\textbf{DeepFool}  & 0.996 & 0.023 $\pm$ 0.026 & 0.005 $\pm$ 0.007 &  0.198 $\pm$ 0.132 & 0.036 $\pm$ 0.032 \\
\cline{2-7}
& \textbf{FGSM}      & 0.588 & 1.122 $\pm$ 4.127 & 0.245 $\pm$ 0.901 & 0.207 $\pm$ 0.154 & 0.035 $\pm$ 0.032	 \\
\hline
\hline

\multirow{3}{*}[-0.3ex]{\rotatebox[origin=c]{90}{Australian}} & \textbf{LowProFool}   & 0.968 & 0.710 $\pm$ 0.530 & \textbf{0.21 $\pm$ 0.141} &  0.374 $\pm$ 0.188 & 0.055 $\pm$ 0.027   \\
\cline{2-7}
&  \textbf{DeepFool}  & 1.0 & 0.50 $\pm$ 0.349 & 0.263 $\pm$ 0.183 &  0.375 $\pm$ 0.189 & 0.055 $\pm$ 0.027\\
\cline{2-7}
& \textbf{FGSM}      & 0 & -- & -- & -- & -- \\
\hline
\hline
\multirow{3}{*}[-0.1ex]{\rotatebox[origin=c]{90}{Lending L.}}   & \textbf{LowProFool}   &0.944 & 0.124 $\pm$ 0.168 & \textbf{0.014 $\pm$ 0.027} & 0.659 $\pm$ 0.207 & 0.062 $\pm$ 0.027  \\
\cline{2-7}
 & \textbf{DeepFool}  &0.996 & 0.107 $\pm$ 0.154 & 0.024 $\pm$ 0.035 &  0.66 $\pm$ 0.209 & 0.062 $\pm$ 0.028 \\
\cline{2-7}
& \textbf{FGSM}      & 0 & -- & -- & -- & -- \\
\hline

\end{tabularx}
\end{table}
\end{center}

\section{Discussion}\label{sec:discussion}


We made the hypothesis in \cref{sec:intro} that perturbations on less relevant features still allow to move sufficiently an example in the feature space to get access to the desired opposed class label. The reported results confirm our hypothesis to some extent. Indeed, we never achieve both 100\% success rate and satisfactory results on the weighted mean norm.
However, aiming for a smaller success rate of 95\%, we reach low perturbation norm ratios between LowProFool and DeepFool. 

This intuitively results from the trade-off introduced with the objective function $g$ defined in \cref{sec:generate}. Take for instance a sample in the set of the original samples $\mathbb X$ that is far from the classifier frontier and for which the closest path to reach the classifier frontier is along highly perceptible feature with regards to the defined feature-importance vector $\bm v$. It is not unexpected that in a fixed number of iterations and given the fact that we highly penalize moves onto highly perceptible features, the sample does not reach the classifier frontier.

Furthermore, the hyperparameters controlling the speed of move as well as the trade-off between class-changing and minimizing the perceptibility value are fitted on the whole validation set. We then believe that these hyperpameters do not generalize enough hence preventing a few outliers from either crossing the classification frontier or minimizing enough their perceptibility value.

To the contrary, the DeepFool method involves no hyperparameters. First, it adapts the speed of the descent by scaling the perturbation with the norm of the difference of the logits towards the target and towards the source. Intuitively, this means that when the sample is far from the classifier's frontier, the perturbation will be large, and will get smaller when the crafted adversarial examples gets close to the frontier.\\
Second, the objective of the DeepFool algorithm is to minimize the $\ell_2$ norm of the perturbation under no constraint regarding the perceptibility of the perturbation. This naturally makes it easier for DeepFool to generate adversarial example compared to LowProFool which also seeks to minimize the norm of the Hadamard product between the feature importance vector $\bm v$ and the perturbation vector $\bm r$
It is hence not surprising for LowProFool not to reach as good success rate results as DeepFool.

Concerning FGSM, it seems that we reach the limits of a method that worked sufficiently well on images but is actually unsuccessful on tabular data.
 

The adversarial examples generated with LowProFool outperform DeepFool's to a great extent in terms of the perceptibility metrics defined in \cref{sec:metrics}. What's more, it did not cost much in terms of lowering the success rate to reach those results. So not only can we generate adversarial examples in the tabular domain using  less important features to the eye of the expert at hand, but we can also achieve great results in terms of imperceptibility of the attack. We believe that the existence of these adversarial examples is directly tied to the discrepancy between the classifier's learned feature importance and the a priori feature importance vector $\bm v$.

The comparison between the weighted mean norm and the distance to the closest neighbors reveals that the generated adversarial examples are very close to their original example. For instance, the average weighted perturbation represents only $5.9\%$ of the average distance to the closest neighbor for the Default Credit dataset. This behaviour is really desirable as it reinforces our belief that the generated adversarial examples are closer to their original sample more than to any other sample hence being the most imperceptible.

About coherence, we were worried that clipping a posteriori would introduce deadlocks. For instance, take a sample for which the gradient of the objective function points towards a unique direction but we restrict any move towards this direction (e.g. negative age). In practice, this behaviour did not show.


\section{Limitations and perspectives}

Our approach constitutes a white-box attack as the attacker has access to the model as well as the whole dataset. While a white-box attack often does not comply with real-world attacks, we believe that it is achievable to perform a black-box attack by using a surrogate model under the assumption to have a minimum number of queries allowed to the oracle.~\citeauthor{blackbox2} showed they could achieve a high rate of adversarial example misclassifications on online deep-learning APIs by creating a substitute model to attack a black-box target
model. To train the substitute model, they synthesize a training set and annotated it by querying the target model for labels~\citep{blackbox2}.~\citeauthor{blackbox} also showed that transferable non-targeted adversarial examples are easy to find~\citep{blackbox}. We seek to investigate this area of research in the future steps. Indeed, black-box approaches would allow us to build attacks that do not rely on gradient information as it is the case for LowProFool.

\section{Conclusion}

In this paper, we have focused our attention on the notion of adversarial examples in the context or tabular data. To the best of our knowledge, there has not been much focus on this data domain, contrary to the images domain that attracted lots of attention. Images and tabular data do not share the same perceptibility concepts and what has been formalized on images cannot be applied on tabular data. We propose a formalization of the notion of perceptibility and coherence of adversarial examples in tabular data, along with a method, LowProFool, to generate such adversarial examples. Proposed metrics show successful results on the ability to fool the model and generate imperceptible adversarial examples. 
Gradient-based methods such as LowProFool show limits when it comes to discrete features. We will investigate new methods to answer this matter.
While our proposition remains subject to challenges, we believe it is a first step towards a more comprehensive and complete view of the field of Adversarial Machine Learning.

\section*{Acknowledgements}

We would like to thank Seyed-Mohsen Moosavi-Dezfooli and Apostolos Modas who provided their comments, discussion and expertise on this research.

\bibliography{lib}

\end{document}